%% file: main.tex
\documentclass[sigconf]{acmart}

\input{section/preamble}

\copyrightyear{2023}
\acmYear{2023}
\setcopyright{acmlicensed}
\acmConference[CIKM '23]{Proceedings of the 32nd ACM International Conference on Information and Knowledge Management}{October 21--25, 2023}{Birmingham, United Kingdom}
\acmBooktitle{Proceedings of the 32nd ACM International Conference on Information and Knowledge Management (CIKM '23), October 21--25, 2023, Birmingham, United Kingdom}
\acmPrice{15.00}
\acmDOI{10.1145/3583780.3615155}
\acmISBN{979-8-4007-0124-5/23/10}

\begin{document}

\title{Toward a Foundation Model for Time Series Data}
\input{section/author}
\input{section/abstract}

\maketitle

\input{section/introduction}
\input{section/model}

\input{section/method}
\input{section/experiment}
\input{section/conclusion}

\bibliographystyle{ACM-Reference-Format}
\balance
\bibliography{section/reference}

\end{document}

%% file: section/preamble.tex
\usepackage{graphicx}

\usepackage{xcolor}

\usepackage{algorithm}
\usepackage[noend]{algpseudocode}

\algrenewcommand\alglinenumber[1]{#1}

\usepackage{multirow}

\usepackage{enumitem}
\setlist[itemize]{leftmargin=*, topsep=.0em, itemsep=0pt, parsep=0pt, partopsep=0pt}
\setlist[enumerate]{leftmargin=*, topsep=.0em, itemsep=0pt, parsep=0pt, partopsep=0pt}

\usepackage{amsthm}
\newtheoremstyle{def_style}
  {.0em}          
  {.0em}          
  {}          
  {}          
  {\bfseries} 
  {.}         
  {.0em}      
  {}          

\theoremstyle{def_style}

\theoremstyle{def_style}

\usepackage{amsmath}
\usepackage{bbm}

\usepackage{collectbox}

\makeatletter
\newcommand{\mybox}{%
    \collectbox{%
        \setlength{\fboxsep}{1pt}%
        \fbox{\BOXCONTENT}%
    }%
}
\makeatother

\usepackage{balance}

%% file: section/author.tex
\author{Chin-Chia Michael Yeh, Xin Dai, Huiyuan Chen, Yan Zheng, Yujie Fan, Audrey Der$^\dag$,\\Vivian Lai, Zhongfang Zhuang, Junpeng Wang, Liang Wang, and Wei Zhang}
\email{{miyeh,xidai,hchen,yazheng,yufan,viv.lai,zzhuang,junpenwa,liawang,wzhan}@visa.com, ader003@ucr.edu}
\affiliation{%
    \institution{Visa Research, University of California, Riverside$^\dag$}
}

\renewcommand{\shortauthors}{Chin-Chia Michael Yeh et al.} 

%% file: section/abstract.tex
\begin{abstract}
A foundation model is a machine learning model trained on a large and diverse set of data, typically using self-supervised learning-based pre-training techniques, that can be adapted to various downstream tasks. 
However, current research on time series pre-training has predominantly focused on models trained exclusively on data from a single domain. 
As a result, these models possess domain-specific knowledge that may not be easily transferable to time series from other domains.
In this paper, we aim to develop an effective time series foundation model by leveraging unlabeled samples from multiple domains. 
To achieve this, we repurposed the publicly available UCR Archive and evaluated four existing self-supervised learning-based pre-training methods, along with a novel method, on the datasets.
We tested these methods using four popular neural network architectures for time series to understand how the pre-training methods interact with different network designs.
Our experimental results show that pre-training improves downstream classification tasks by enhancing the convergence of the fine-tuning process. 
Furthermore, we found that the proposed pre-training method, when combined with the Transformer, outperforms the alternatives.
The proposed method outperforms or achieves equal performance compared to the second best method in $\sim93$\% of downstream tasks.
\end{abstract}

\ccsdesc[500]{Computing methodologies~Neural networks}
\ccsdesc[500]{Computing methodologies~Temporal reasoning}

\keywords{time series, self-supervised learning, foundation model}

%% file: section/introduction.tex
\section{Introduction}
\label{sec:introduction}
Foundation models are a type of machine learning model that are trained through self-supervised learning on large-scale, multi-domain datasets and can be fine-tuned for a wide range of downstream tasks~\cite{bommasani2021opportunities}.
Although extensively studied in natural language processing and computer vision communities~\cite{devlin2018bert,bommasani2021opportunities,yuan2021florence,radford2021learning}, there has been surprisingly little work focused on building foundation models for time series data.
Several papers have explored self-supervised learning with time series data~\cite{tang2020exploring,zhang2022self,yue2022ts2vec,wickstrom2022mixing}, but none have tested their methods on a pre-training dataset consisting of time series from multiple domains.
In other words, while these methods may provide valid solutions for training foundation models for time series data, experiments involving multi-domain pre-training datasets have not yet been conducted.

To investigate this research problem, we utilized the publicly available time series dataset archive, the UCR Archive~\cite{dau2019ucr}, and repurposed it for our experimental setup as illustrated in Fig.~\ref{fig:setup}.
We compiled a pre-training set comprising time series from all domains in the UCR Archive, ranging from power consumption to heartbeats and food spectrographs.
Furthermore, we generated multiple mutually exclusive training, validation, and test sets for downstream classification tasks.
Using the repurposed UCR Archive, we aimed to address the following three research questions:
\begin{itemize}
\item Can pre-training a foundation model on a multi-domain dataset lead to improved performance on downstream single-domain classification tasks after fine-tuning?
\item What neural network architectures are most effective for the foundation model?
\item What is the most effective self-supervised learning method for building the foundation model?
\end{itemize}

\input{insert/fig_setup}

To address these research questions, we built foundation models using four widely used neural network architectures (Section~\ref{sec:model}), and trained them using four existing and one novel self-supervised learning methods (Section~\ref{sec:method}).
In total, we tested 20 different configurations for building foundation models.
Our experimental results showed that pre-training a foundation model with a multi-domain dataset can improve the performance of downstream single-domain classification tasks after fine-tuning.
One possible reason for the improved performance using a foundation model is that it has a smoother convergence curve (i.e., better convergence) when fine-tuned compared to training a model from scratch.
Furthermore, we found that the best-performing foundation model was achieved by combining the proposed self-supervised learning method with the Transformer architecture.
Our contributions in this study are:
\begin{itemize}
\item We describe a set of steps to repurpose the UCR Archive for the study of building time series foundation models.
\item We demonstrate that utilizing foundation models trained with time series from multiple domains can indeed benefit downstream classification tasks.
\item We observe that the convergence curve of the foundation model during the fine-tuning stage is much smoother compared to training a model from scratch, which helps the pre-trained models achieve better performance.
\item We propose an effective learning method for training time series foundation models that outperforms alternative methods.
\end{itemize}

%% file: insert/fig_setup.tex
\begin{figure}[ht]
\centerline{
\includegraphics[width=0.85\linewidth]{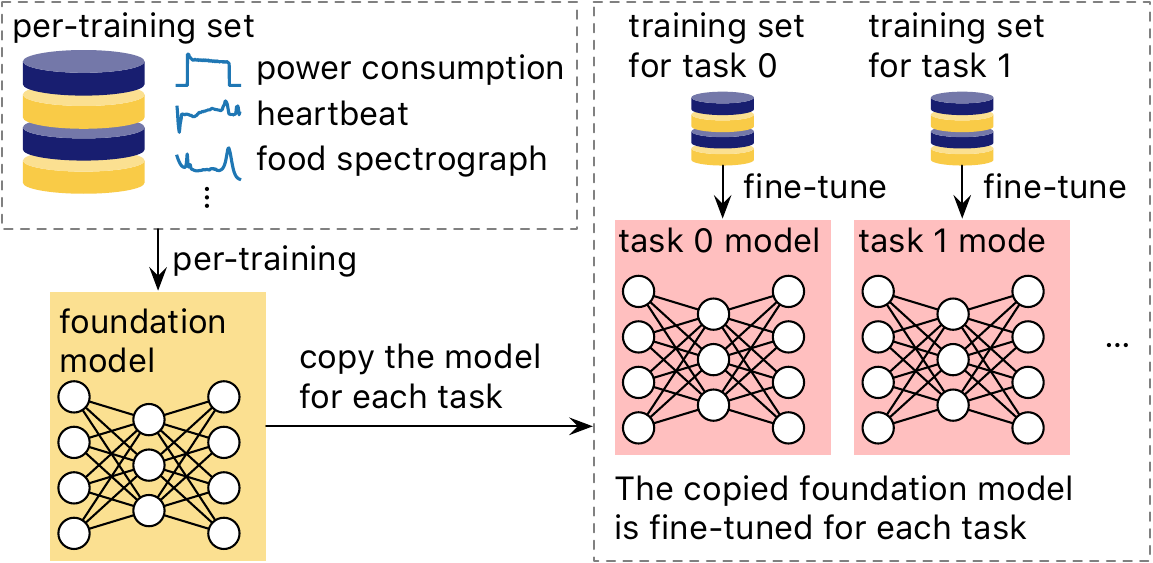}
}
\caption{
The foundation model is initially trained on a large dataset of time series from various domains, and is then fine-tuned using a smaller training set for each downstream task.
}
\label{fig:setup}
\end{figure}

%% file: section/model.tex
\section{Model Architecture}
\label{sec:model}
We explored four different neural network architectures as the backbone model for the foundation models:

\noindent \textbf{LSTM}: 
The Long Short-Term Memory Network (LSTM) is a widely-used type of Recurrent Neural Network (RNN) for modeling sequential data~\cite{hochreiter1997long,lim2021time,zhou2021informer}. 
We use the design illustrated in Fig.~\ref{fig:model}.c.
In the figure, we use the notation \mybox{\texttt{1D conv,7/2,1${\to}$64}} to denote a $1D$ convolutional layer with a filter size of 7, a stride size of 2, an input dimension of 1, and an output dimension of 64. 
Similarly, we use the notation \mybox{\texttt{bi-RNN,64${\to}$64}} to denote a bidirectional RNN layer with an input dimension of 64 and an output dimension of 64.
In this case, the two \texttt{bi-RNN} layers are bidirectional LSTM layers.
Finally, we use the notation \mybox{\texttt{linear,64${\to}$64}} to denote a linear layer with an input dimension of 64 and an output dimension of 64.

\noindent \textbf{GRU}: 
The Gated Recurrent Unit Network (GRU) is another popular type of RNN architecture widely used for modeling sequential data~\cite{cho2014properties,lim2021time,zhou2021informer}. 
We adopt the design depicted in Fig.~\ref{fig:model}.c with two bidirectional GRU layers.

\noindent \textbf{ResNet}: 
The Residual Network (ResNet) is a time series classification model that takes inspiration from the success of ResNet in computer vision~\cite{he2016deep,wang2017time}. 
Extensive evaluations by~\cite{ismail2019deep} has demonstrated that ResNet is one of the strongest models for time series classification. 
Our design, as shown in Fig.~\ref{fig:model}.b, is based on the design proposed by~\cite{wang2017time}. 
We use the notation \mybox{\texttt{block, 64$\to$64}} to denote a residual block (Figure\ref{fig:model}.a) with an input dimension of 64 and an output dimension of 64.

\noindent \textbf{Transformer}: 
The Transformer (XFMR) is a widely used alternative to RNNs for sequence modeling~\cite{vaswani2017attention,li2019enhancing,zhou2021informer,lim2021time,chen2022denoising}. 
The architecture we used is shown in Fig.~\ref{fig:model}.d. 
We used fixed positional encoding, following~\cite{vaswani2017attention}, and included a special token \texttt{[start]} to learn the representation of the entire time series. 
Our Transformer architecture consists of four encoder layers, each denoted as \mybox{\texttt{XFMR,8,64${\to}$64}} where the number of heads is 8, the input dimension is 64, and the output dimension is 64.

We use layer normalization~\cite{ba2016layer} for all normalization layers, as it is both effective and widely used in modeling sequential data~\cite{ba2016layer,vaswani2017attention}.

\input{insert/fig_model}

%% file: insert/fig_model.tex
\begin{figure}[t]
\centerline{
\includegraphics[width=0.85\linewidth]{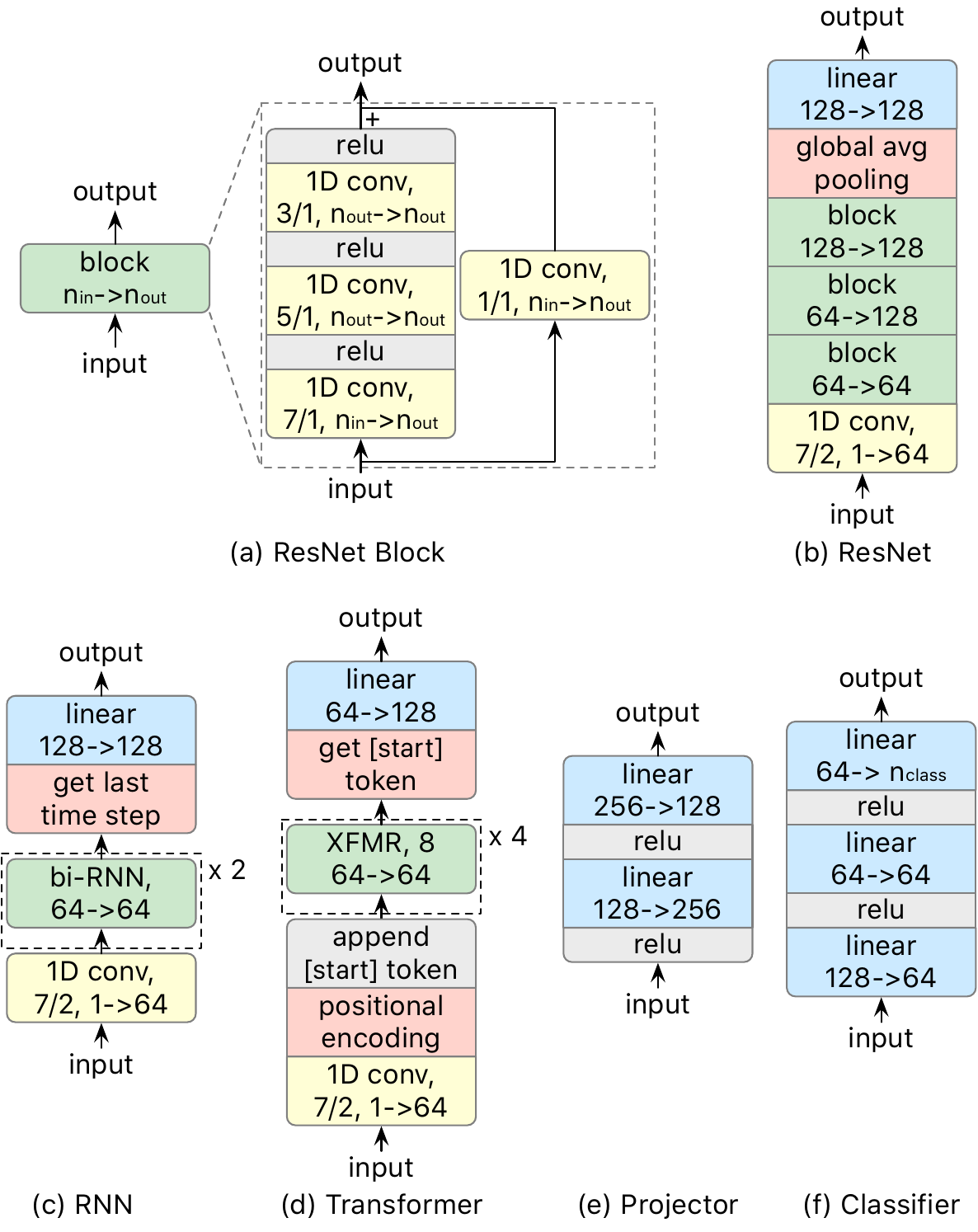}
}
\caption{
The designs for ResNet, RNN, and Transformer are illustrated in Fig.~\ref{fig:model}.a to d. 
The projector and classifier are used for pre-training and fine-tuning.
}
\label{fig:model}
\end{figure}

%% file: section/method.tex
\section{Pre-training Method}
\label{sec:method}
We evaluated a total of five pre-training methods. 
The first four methods listed below serve as the baseline methods, while the fifth method is the proposed method.

\noindent \textbf{SimCLR}:
The SimCLR method was originally proposed as a self-supervised pre-training method based on contrastive learning for computer vision~\cite{chen2020simple}, and was later extended to human activity time series by~\cite{tang2020exploring}. 
The pre-training procedure for SimCLR involves a batch of time series $X$ that undergoes a sequence of randomized data augmentation methods to generate two augmented batches $X_0$ and $X_1$. 
The data augmentation methods used are random scaling and negation, following the open-sourced implementation~\cite{zhang2022self,contributions2023tfcgithub}. 
Both $X_0$ and $X_1$ are then processed by the backbone model (Section~\ref{sec:model}) and the projector (Fig.~\ref{fig:model}.e); the output feature vectors are denoted as $H_0$ and $H_1$, respectively.

\sloppy
Next, the NT-Xent loss function~\cite{chen2020simple,tang2020exploring} is used to compute the loss. 
If $h_i \in H_0$ and $h_j \in H_1$ are features extracted from the augmented versions of the same time series in $X$, the loss for the positive pair $(h_i, h_j)$ is computed as follows:
\begin{equation}
\footnotesize
\mathcal{L} = - \log \frac{\text{exp} \left(\text{sim}(h_i, h_j)/\tau\right)}{\sum_{h_k \in H_0 + H_1} \mathbbm{1}_{[h_k \neq h_i \& h_k \neq h_j]} \text{exp}\left(\text{sim}(h_i, h_k)/\tau\right)}
\end{equation}
Here, $\text{sim}(\cdot, \cdot)$ is a function that computes the cosine similarity between the input vectors, $h_k$ is a feature vector from $H_0$ or $H_1$ that is not $h_i$ nor $h_j$, and $\tau$ is the temperature parameter.
Both the backbone model and the projector are optimized using the NT-Xent loss. 
To fine-tune the model for the downstream classification task, we add a classifier model (Fig.~\ref{fig:model}.f) on top of the projector as shown in Fig.~\ref{fig:pretrain}.a, and update the backbone, projector, and classifier model using the cross-entropy loss similar to the standard supervised learning setting~\cite{ismail2019deep}.

\noindent \textbf{TS2Vec}:
The TS2Vec method is another pre-training method that utilizes contrastive learning~\cite{yue2022ts2vec}. 
What differentiates the TS2Vec method from SimCLR is that: 1) the positive pairs are generated using the contextual consistency strategy and 2) the contrastive loss is calculated in a hierarchical fashion. 
The contextual consistency strategy generates positive pairs by randomly cropping two overlapping subsequences from a time series. 
The TS2Vec method computes the contrastive loss under multiple levels of time granularity. 
At each time granularity, two contrastive losses are computed: temporal contrastive loss and instance-wise contrastive loss. 
Both are modified NT-Xent losses where the former helps the model learn discriminative representations over time, and the latter helps the model learn discriminative representations over samples.

\sloppy
Note that TS2Vec requires the backbone model to generate representation for multiple time steps. 
To achieve this, we modify the RNN-based models (Fig.~\ref{fig:model}.c) by removing the \mybox{\texttt{get last time step}} block and apply the subsequent linear layer and projector model independently at each time step to generate the per-time-step representation. 
For ResNet and Transformer (Fig.~\ref{fig:model}.b and Fig.~\ref{fig:model}.d), we remove the \mybox{\texttt{global avg pooling}} and \mybox{\texttt{get [start] token}} block, respectively, and apply the subsequent linear layer and projector model independently at each time step to generate the per-time-step representation. 
Because the first layer for all the backbone models is a $1D$ convolutional layer with a stride size of 2, the length of the output sequence is halved from the input time series. 
To fine-tune the model for classification, we first re-add the removed blocks back to the model, add a classifier model (Fig.~\ref{fig:model}.f) on top of the projector as shown in Fig.~\ref{fig:pretrain}.a, and update the backbone, projector, and classifier model using the cross-entropy loss.

\noindent \textbf{MixingUp}:
The MixingUp method was also developed based on the idea of contrastive learning~\cite{wickstrom2022mixing}. 
Given a pair of time series $(x_i, x_j)$, it generates a mixed time series $x_k$ by computing $\lambda x_i + (1-\lambda) x_j$, where $\lambda$ is the mixing parameter that determines the contribution of $x_i$ and $x_j$ in $x_k$. 
The mixing parameter is randomly drawn from a beta distribution.
The self-supervised task for pre-training is to predict $\lambda$ from the representations associated with $x_i$, $x_j$, and $x_k$. 
The representations are generated by processing $x_i$, $x_j$, and $x_k$ with the backbone and projector model. 
The loss function adopted by~\cite{wickstrom2022mixing} is a modified version of the NT-Xent loss.
Similar to the previous two methods, fine-tuning for the classification task is achieved by adding a classifier model on top of the projector (Fig.~\ref{fig:pretrain}.a) and updating the backbone, projector, and classifier model using the cross-entropy loss.

\input{insert/fig_pretrain}

\noindent \textbf{TF-C}:
The TF-C method extends the idea of contrastive learning to the frequency domain~\cite{zhang2022self}. 
Based on the open-source implementation~\cite{zhang2022self,contributions2023tfcgithub}, the self-supervised learning procedure is as follows:
The method involves transforming a time series from the time domain to the frequency domain using the Fast Fourier transform (FFT). 
Positive pairs are generated through augmentation functions applied to the time series in both the time and frequency domains. 
TF-C employs the jittering technique to augment the time series in the time domain and the technique of adding/removing frequency components in the frequency domain. 
The input time series are processed in both the time and frequency domains using their respective backbone models and projectors.
The self-supervised learning loss is calculated using a modified NT-Xent loss function, which involves intermediate representations from both the time and frequency domains.
Since the TF-C method uses two backbone models and two projector models, the width of these models is halved so that it has the same number of parameters as the other methods. 
The structure of the model for fine-tuning is shown in Fig.~\ref{fig:pretrain}.b, where the output of the projector in time and frequency domain is concatenated before being fed into the classification model. 
The classification fine-tuning is once again achieved with cross-entropy loss.

\noindent \textbf{TimeCLR}:
This is our proposed contrastive learning pre-training method, building on the simplest existing method, SimCLR~\cite{tang2020exploring}.
The first enhancement we introduce is the incorporation of additional time series data augmentation techniques based on recent survey papers~\cite{wen2020time,iwana2021empirical}.
Our method employs the following data augmentation techniques: jittering, smoothing, magnitude warping, time warping, circular shifting, adding slope, adding spike, adding step, masking, and cropping. 
These additional techniques enhance the pre-training process by allowing the model to learn more invariance properties. 
For instance, jittering and smoothing aid the model in being invariant to Gaussian noise, and magnitude and time warping help the model become warping invariant in both aspects. 
Circular shifting, adding slope/spike/step, and masking make the model invariant to corresponding noise types, while cropping helps the model learn contextual consistency similar to TS2Vec~\cite{yue2022ts2vec}.


Our second enhancement involves using a single augmentation function for generating positive pairs instead of using all augmentation functions. 
This is due to the fact that we incorporate more data augmentation functions in our method compared to the existing SimCLR implementation~\cite{zhang2022self,contributions2023tfcgithub}.
Applying all augmentation methods to the input time series could result in the augmented time series bearing little visual similarity to the original. 
For example, when we apply all the data augmentation methods to a simple pattern \input{insert/fig_ex_org}~, the resulting time series \input{insert/fig_ex_aug} loses many of the properties of the original time series.
It is possible to achieve better performance by using more than one augmentation when generating positive pairs, but we left this exploration for future work.
Apart from these two enhancements, both the pre-training and fine-tuning processes are identical to those of SimCLR. 
For fine-tuning, we add a classifier model on top of the projector (as shown in Fig.~\ref{fig:pretrain}.a) and update the backbone, projector, and classifier model using the cross-entropy loss. 

%% file: insert/fig_pretrain.tex
\begin{figure}[ht]
\centerline{
\includegraphics[width=0.75\linewidth]{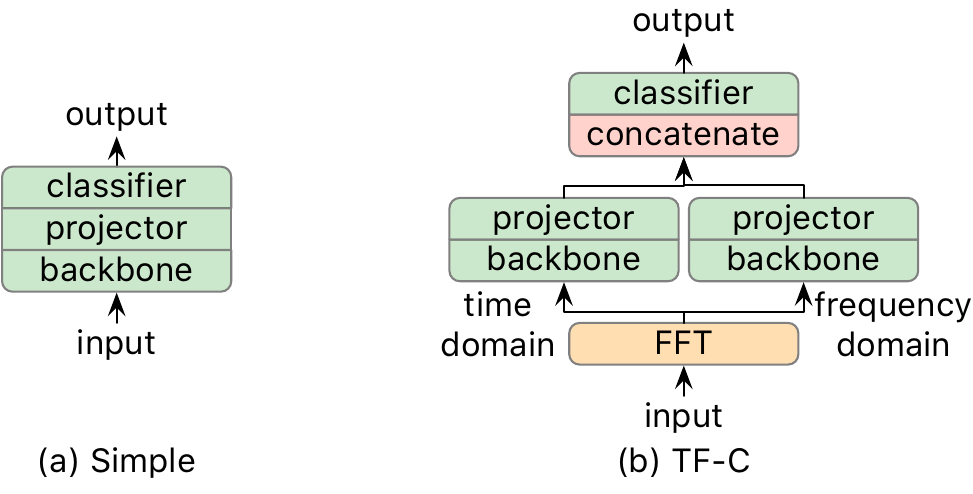}
}
\caption{
There are two ways to use backbone models for pre-training and fine-tuning.
Fig.~\ref{fig:pretrain}.a is used for all methods except TF-C, which uses Fig.~\ref{fig:pretrain}.b.
}
\label{fig:pretrain}
\end{figure}

%% file: insert/fig_ex_org.tex
(\includegraphics[height=0.9em]{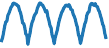})

%% file: insert/fig_ex_aug.tex
(\includegraphics[height=0.9em]{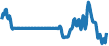})

%% file: section/experiment.tex
\section{Experiment}
\label{sec:experiment}
To ensure the reproducibility of our experiments, we have created a website~\cite{supplementary} that contains supplementary materials, including hyper-parameter settings, source code, and complete experimental results.
We first explain how we repurposed the datasets in the UCR Archive~\cite{dau2019ucr}. 
We combined the original training set with the test set for each of the 128 datasets in the UCR Archive. 
Next, we created a pre-training set by randomly extracting 50\% of the samples from each of the 128 datasets in the UCR Archive. 
In other words, the pre-training set contains time series samples from all 128 datasets.
We then split the remaining time series samples in each of the 128 datasets into training, validation, and test sets with a ratio of 3:1:1. 
As a result, we have a single pre-training set, 128 training sets, 128 validation sets, and 128 test sets. 
It is important to note that all split sets are mutually exclusive, and there is no label information available in the pre-training set.

We combine the five pre-training methods (SimCLR, TS2Vec, MixingUp, TF-C, and TimeCLR) with the four neural network architectures (LSTM, GRU, ResNet, and Transformer), resulting in 20 different configurations for the pre-training methods.
First, we train the foundation model using each configuration with the pre-training set.
Next, we copy each trained foundation model for each of the 128 downstream tasks, where we train with the corresponding training set for the task.
We perform model selection\footnotemark{} with the task-specific validation set and measure the performance (i.e., accuracy) with the associated test set.

\footnotetext{The model selection process selects the model from the best epoch and initialization strategy (i.e., random or pre-train) based on validation accuracy.}

We also include a \textit{no pre-training} baseline for each network architecture. 
For this baseline, we perform model training, model selection, and evaluation using only the task-specific dataset. 
In addition to the neural network-based methods, we also include one-nearest-neighbor with Euclidean Distance (ED) and Dynamic Time Warping distance (DTW) as baselines in our experiment. 
ED and DTW are considered simple yet effective baselines for time series classification problems~\cite{rakthanmanon2012searching,bagnall2017great,dau2019ucr}. 
The experimental results across 128 downstream tasks are summarized in Table~\ref{tab:summary}.
We report the average rank among the tested methods in the table.

\input{insert/tab_summary}

When we consider the method without any pre-training, we can see that ResNet and Transformer outperform the other methods. 
The superior performance of ResNet is not surprising as the results are in agreement with~\cite{ismail2019deep}. 
The success of Transformer is also expected, as Transformers have achieved state-of-the-art performance in natural language processing~\cite{vaswani2017attention,liu2019roberta,brown2020language}, computer vision~\cite{dosovitskiy2020image,chen2021empirical,han2022survey}, and other time series problems~\cite{zerveas2021transformer,nie2022time,wen2022transformers}. 
On the other hand, the performance of RNN-based models (LSTM and GRU) is worse than simple baselines like ED and DTW.

When comparing the performance of various pre-trained models with their non-pre-trained counterparts, we can observe that the foundation model is beneficial in 14 out of 20 configurations. 
In particular, aside from SimCLR (which was designed specifically for human activity time series~\cite{tang2020exploring}), the performance of RNN-based models improved greatly relative to the configurations with ResNet or Transformer, such that the RNN-based models had comparable performance with the other two model architectures.
Generally speaking, the foundation model can improve the performance of downstream tasks after fine-tuning. 
Additionally, when comparing the proposed TimeCLR method with other pre-training methods, we can see that the proposed TimeCLR achieves the best performance regardless of the type of architecture used for the backbone model. 
The overall best performance is achieved by combining the TimeCLR method with the Transformer backbone model.

In Fig.~\ref{fig:converge}, we compare the convergence curves of non-pre-trained and TimeCLR pre-trained models on two datasets. 
The ArrowHead dataset is considered ``harder" to converge compared to CBF.
Pre-trained models on the harder dataset exhibit smoother convergence curves during fine-tuning compared to training from scratch, despite both experiments using the same model architecture and optimization settings.
In contrast, for the easier dataset, both models behave similarly, and either model can be trained to achieve good performance. 
This difference in convergence for harder datasets is likely the main reason why the foundation model improves the overall performance of downstream tasks.

\input{insert/fig_converge}


%% file: insert/tab_summary.tex
\begin{table}[ht]
\centering
\caption{
The average ranks of different methods across 128 datasets in UCR Archive are reported, with lower ranks indicating better performance.
XFMR means Transformer. 
For each backbone model, we highlight the best performance in bold and the second best performance with an underline.
} 
\label{tab:summary}
\footnotesize
\begin{tabular}{l||cc|cccc}
Average rank ($\downarrow$) & ED & DTW & LSTM & GRU & ResNet & XFMR \\ \hline \hline
No pre-train & 15.14 & 14.08  & 21.33 & 20.96 & 11.16 & 11.38 \\ \hline
SimCLR & - & - & 21.38 & 20.98 & 11.12 & 11.77 \\ 
TS2Vec & - & - & \underline{13.71} & 16.07 & 10.56 & \underline{11.37} \\ 
MixingUp & - & - & 14.18 & \underline{12.16} & 11.19 & 11.38 \\ 
TF-C & - & - & 14.53 & 13.24 & \underline{10.55} & 11.48 \\ \hline
TimeCLR  & -  & - & \textbf{12.43} & \textbf{9.51} & \textbf{10.04}  & \textbf{9.30} \\ 
\end{tabular}
\end{table}

%% file: insert/fig_converge.tex
\begin{figure}[ht]
\centerline{
\includegraphics[width=0.95\linewidth]{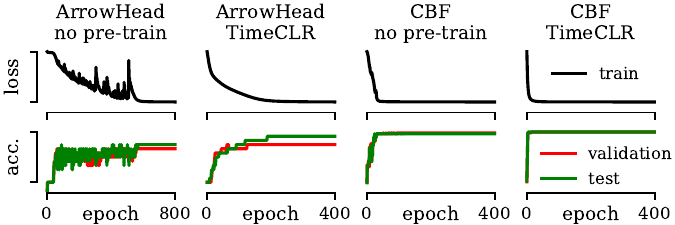}
}
\caption{
The convergence curves of both the non-pre-trained ResNet model and TimeCLR pre-trained ResNet model on the ArrowHead and CBF dataset. 
}
\label{fig:converge}
\end{figure}

%% file: section/conclusion.tex
\section{Conclusion}
\label{sec:conclusion}
In this paper, we have demonstrated the benefits of using self-supervised learning to train a foundation model for time series data from multiple domains. 
The proposed TimeCLR method outperformed the alternatives when combined with transformer models. 
For future work, it would be interesting to explore other ways of combining the benefits of different pre-training methods~\cite{tang2020exploring,zhang2022self,yue2022ts2vec,wickstrom2022mixing}, the possibility of using the data compression idea in self-supervised learning~\cite{tan2020learning,qin2020ghashing,yeh2022embedding,yeh2022error}, and testing alternative backbone model like the ResNet $2D$~\cite{yeh2023efficient,yeh2023multitask}.